\documentclass[11pt]{article}

\usepackage{microtype} 
\usepackage{booktabs}  
\usepackage{url}  

\usepackage{graphicx}
\usepackage{wrapfig}
\usepackage[enable]{easy-todo}

\usepackage{amsmath}
\usepackage{amsthm}

\usepackage[final,shortpaper]{automl}
\usepackage{natbib}
\title{Surrogate Benchmarks for Model Merging Optimization}

\author[1]{\nameemail{Rio Akizuki}{akizuki-rio-pd@ynu.jp}}
\author[1]{\nameemail{Yuya Kudo}{kudo-yuya-dr@ynu.jp}}
\author[1]{\nameemail{Nozomu Yoshinari}{yoshinari-nozomu-ry@ynu.jp}}
\author[1]{\nameemail{Yoichi Hirose}{hirose-youichi-kc@ynu.jp}}
\author[1]{\nameemail{Toshiyuki Nishimoto}{nishimoto-toshiyuki-gf@ynu.jp}}
\author[1]{\nameemail{Kento Uchida}{uchida-kento-fz@ynu.ac.jp}}
\author[1]{\nameemail{Shinichi Shirakawa}{shirakawa-shinichi-bg@ynu.ac.jp}}

\affil[1]{Yokohama National University}

\hypersetup{%
  pdfauthor={}, 
  pdftitle={},
  pdfsubject={},
  pdfkeywords={}
}

\begin{document}

\maketitle

\begin{abstract}
Model merging techniques aim to integrate the abilities of multiple models into a single model. Most model merging techniques have hyperparameters, and their setting affects the performance of the merged model. Because several existing works show that tuning hyperparameters in model merging can enhance the merging outcome, developing hyperparameter optimization algorithms for model merging is a promising direction. However, its optimization process is computationally expensive, particularly in merging LLMs. In this work, we develop surrogate benchmarks for optimization of the merging hyperparameters to realize algorithm development and performance comparison at low cost. We define two search spaces and collect data samples to construct surrogate models to predict the performance of a merged model from a hyperparameter. We demonstrate that our benchmarks can predict the performance of merged models well and simulate optimization algorithm behaviors.
\end{abstract}

\section{Introduction}
Model merging~\citep{Yang2024} is a promising approach to build a powerful single model from multiple separate models without accessing large datasets and requiring expensive computations. Model merging techniques have succeeded in enhancing the ability of large language models (LLMs) by merging multiple models fine-tuned by different downstream tasks. Most model merging techniques have hyperparameters to set before model merging.
Because these hyperparameters affect the performance of merged models, tuning such hyperparameters can maximize the model merging capabilities. \citet{Akiba2025} proposed the evolutionary model merging that optimizes hyperparameters in model merging using an evolutionary algorithm. They used the covariance matrix adaptation evolution strategy (CMA-ES)~\citep{Hansen2006cma,Hansen2023cmatutorial} and succeeded in finding the superior merging hyperparameters and building high-performance models. 

We call the approach for optimizing hyperparameters in model merging techniques \textit{model merging optimization}. Model merging optimization is a kind of automated machine learning (AutoML) task. The literature on evolutionary model merging~\citep{Akiba2025} shows the effectiveness and potential of the model merging optimization approach using the standard CMA-ES. Therefore, the development of sophisticated model merging optimization algorithms is a promising direction in the AutoML community. However, the computational cost of model merging optimization is relatively high, as with hyperparameter optimization and neural architecture search (NAS).\footnote{For example, the evaluation of a merging configuration for DARE-TIES~\citep{Akiba2025} took about five minutes on an NVIDIA A100 (40GB) in our implementation. Although there is work~\citep{Mencattini2025} to reduce the computational cost of evolutionary model merging by reducing the evaluation dataset and using a performance estimator, it still requires about 3.5 minutes on a single NVIDIA 4090 with 24GB of VRAM. Therefore, when yielding 1,000 merge evaluations, one run of the optimization algorithm requires more than 58 hours.} This large computational load will burden the development of new model merging optimization algorithms and increase the cost of algorithm comparison.

The benchmarks for hyperparameter optimization and NAS~\citep{Eggensperger2015,Ying2019nasbench101,Dong2020nasbench201,Hirose2021nashpo2,Zela2022nasbench301} greatly contribute to algorithm development and evaluation. There are two types of benchmarks: tabular and surrogate benchmarks, where tabular benchmarks provide table lookup for hyperparameter settings and their evaluations through prior exhaustive search, and surrogate benchmarks construct a regression model that returns the performance values from a hyperparameter setting using a sampled actual evaluation dataset.
While tabular benchmarks can provide exact hyperparameter evaluations, creating them in continuous or large search spaces is intractable. Surrogate benchmarks can be constructed even in a continuous and large search space, while the provided hyperparameter evaluation values are predicted by a certain model.
Referring to the success of benchmarks for hyperparameter optimization and NAS, the benchmark for model merging optimization will also be essential for further algorithm development, which will enable us to compare model merging optimization methods at low cost and realize a fair and reproducible comparison.

\citet{He2025} have proposed a benchmark suite for evaluating model merging techniques, which provides standardized fine-tuning models and evaluation protocols. However, evaluating the performance of merged models requires model merging computation and LLM inferences. Therefore, it cannot reduce the cost of model merging optimization.

In this work, we construct a surrogate benchmark to significantly reduce the evaluation cost of model merging optimization. To our knowledge, this is the first surrogate benchmark for model merging optimization.
We collect the paired data of the hyperparameter for model merging and its evaluation values, and construct the surrogate model that predicts the evaluation values from a given hyperparameter.
We evaluate the proposed surrogate benchmark for model merging optimization, termed \texttt{SMM-Bench}, and demonstrate the use of our surrogate benchmark. The code of \texttt{SMM-Bench} is available at \url{https://github.com/shiralab/SMM-Bench}.

\section{Surrogate Model Merging Benchmark (SMM-Bench)}
\citet{Akiba2025} optimized model merging configurations in two model merging settings: parameter space (PS) and data flow space (DFS) merging. For PS merging, where the parameters of the multiple source models are aggregated, the continuous hyperparameters in DARE-TIES~\citep{Yadav2024,Yu2024} were optimized. For DFS merging, the merged model is constructed by stacking source models' layers, and binary variables for the choice of layers and continuous variables for input scaling were optimized. 
Because the hyperparameters in model merging contain continuous variables and can be high-dimension, a surrogate benchmark is a reasonable choice. We introduce surrogate benchmarks for PS and DFS merging as \texttt{SMM-Bench-PS} and \texttt{SMM-Bench-DFS}.

We use Japanese mathematics as a task to evaluate merged models as in \citet{Akiba2025}. The datasets, \texttt{gsm8k-ja-test\_250\_1319} (denoting \texttt{gsm8k-ja} in short)~\citep{cobbe2021training,Akiba2025} and the Japanese test set of \texttt{MGSM}~\citep{shi2023language}, are used to calculate objective values in optimization and final test score, respectively. To evaluate the ability to solve mathematical tasks and provide answers in Japanese, we calculate accuracy, defined as the ratio of correct answers and reasoning texts in Japanese. This evaluation protocol is the same as in \citet{Akiba2025}.

\subsection{SMM-Bench-PS}
\paragraph{Search Space Design}
Tuning the hyperparameters for each layer, called layer-wise merging, has the potential for performance improvement~\citep{Yang2024adamerging}, while it increases the number of hyperparameters to be tuned.
We focus on layer-wise merging using two source models and the simple merging method of task arithmetic~\citep{Ilharco2023}. We use \texttt{shisa-gamma-7b-v1}\footnote{\url{https://huggingface.co/augmxnt/shisa-gamma-7b-v1}} and \texttt{WizardMath-7B-V1.1}~\citep{luo2025wizardmath} as source LLMs for PS merging. These LLMs are fine-tuned from \texttt{Mistral-7B-v0.1}~\citep{jiang2023mistral} and consist of 32 layers. Task arithmetic has a hyperparameter \texttt{weight}, the weighting factor of a task vector. Considering layer-wise merging, different hyperparameters can be specified for each source model's layer, resulting in 64 design variables. We restrict the search space to $[0, 1]^{64}$.

\paragraph{Data Collection}
We created a dataset of merging hyperparameters and their evaluation values on \texttt{gsm8k-ja} and \texttt{MGSM} based on three strategies: random sampling, CMA-ES, and the tree-structured Parzen estimator (TPE)~\citep{Bergstra2011tpe}. Random sampling collected 64,000 data uniformly at random. We ran CMA-ES 13 times for 188 generations with the default hyperparameter setting and TPE implemented in Optuna~\citep{Akiba2019} 12 times for 300 iterations with a batch size of 8. In addition, we uniformly randomly sampled 1,500 data in the model-wise merging setting, i.e., sampling from a two-dimensional subspace. We collected 133,404 data points in total.

\begin{wraptable}{r}[0mm]{0.4\linewidth}
  \caption{The predictive performance of surrogate models.}
  \label{tab:r2-tau}
  \centering
  \begin{tabular}{llrr}
    \toprule
    & Dataset & $R^2$ & KT \\
    \midrule
    PS & \texttt{gsm8k-ja} & 0.950 & 0.883 \\
     & \texttt{MGSM} & 0.921 & 0.791 \\
    \midrule
    DFS & \texttt{gsm8k-ja} & 0.962 & 0.863 \\
     & \texttt{MGSM} & 0.957 & 0.839 \\
    \bottomrule
  \end{tabular}
\end{wraptable}

\paragraph{Surrogate Model Construction}
We split the dataset into training and test sets with a 9:1 ratio. Separate surrogate models predicting the \texttt{gsm8k-ja} and \texttt{MGSM} scores from the merging hyperparameters are trained using the training set. We used LightGBM~\citep{Ke2017LightGBM} as the surrogate model because it exhibited good performance on the surrogate NAS benchmark~\citep{Zela2022nasbench301}. The hyperparameters in LightGBM are optimized using Optuna with five-fold cross-validation. The best-performing cross-validated models are used as our surrogate model by averaging the five models' outputs. Table \ref{tab:r2-tau} shows the $R^2$ score and the Kendall's Tau coefficient (KT) of the predictions made by the surrogate models for the test set. We observe that our surrogate models achieved good prediction performance for both the \texttt{gsm8k-ja} and \texttt{MGSM} scores.

\subsection{SMM-Bench-DFS}

\paragraph{Search Space Design}
Referring to \citet{Akiba2025}, we define a mixed category-continuous search space for DFS merging. We use \texttt{EvoLLM-JP-v1-7B},\footnote{\url{https://huggingface.co/SakanaAI/EvoLLM-JP-v1-7B}} called \texttt{model A}, and \texttt{shisa-gamma-7b-v1}, called \texttt{model B}, as the source models. These source models consist of 32 layers. We construct a merged model by inserting up to 32 layers selected from \texttt{model A} and \texttt{model B} between the 31st and 32nd layers of \texttt{model A}. The $i$-th layer of the 32 potential inserted layers is selected from three options: the $i$-th layer of \texttt{model A}, the $i$-th layer of \texttt{model B}, and without insertion. This layer insertion is determined by a 32-dimensional categorical variable with three categories. In addition, we introduce layer input scaling factors as hyperparameters to mitigate the input distribution shift~\citep{Akiba2025}. We fix the first layer's scaling of the merged model to 1.0 and treat those for the other 63 layers, including potentially inserted layers, as hyperparameters. The range of scaling factors is $[0.4, 1.5]$. As a result, the merging hyperparameters consist of 32 categorical variables and 63 continuous variables.

\paragraph{Data Collection}
We totally collected 40,913 data points, where 22,286 data were sampled uniformly at random from the search space. We ran CatCMA~\citep{Hamano2024,nomura2024cmaes} three times for 177 generations with the default setting and TPE four times for 300 iterations with a batch size of 8.

\paragraph{Surrogate Model Construction}
The surrogate models are trained in the same procedure as \texttt{SMM-Bench-PS}. The $R^2$ score and Kendall's Tau coefficient (KT) for the test set are also displayed in Table \ref{tab:r2-tau}. Our surrogate models achieved high predictive performance, while the dataset size was smaller than that of \texttt{SMM-Bench-PS}.

\section{Evaluation of SMM-Bench}
\begin{figure}[tb]
  \begin{subfigure}[t]{0.99\linewidth}
    \centering
    \includegraphics[width=0.85\linewidth]{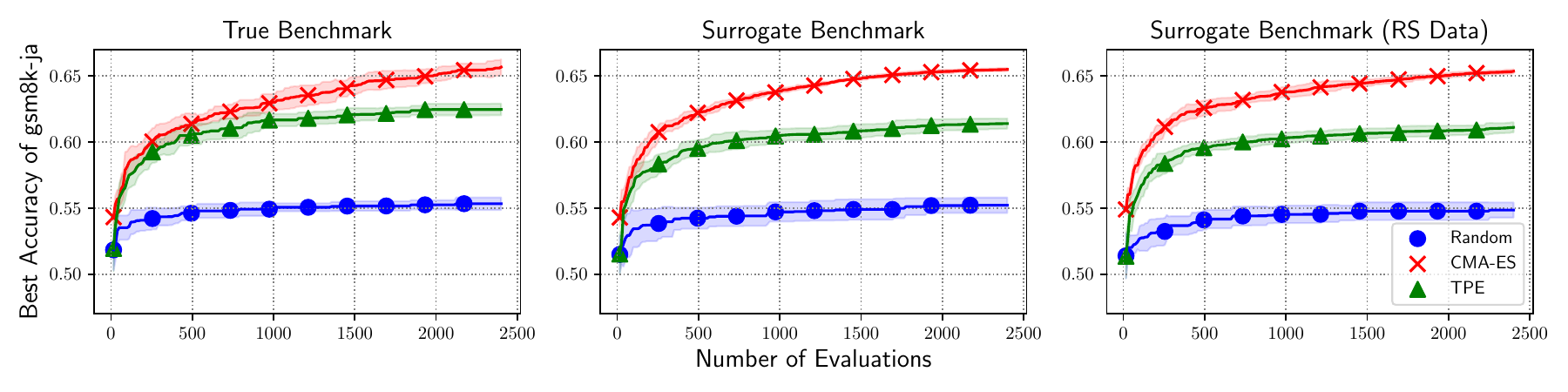}
    \vspace{-0.25cm}
    \caption{Search space for PS merging (\texttt{SMM-Bench-PS})}
    \label{fig:ps_traj}
  \end{subfigure}
  \begin{subfigure}[t]{0.99\linewidth}
    \centering
    \includegraphics[width=0.85\linewidth]{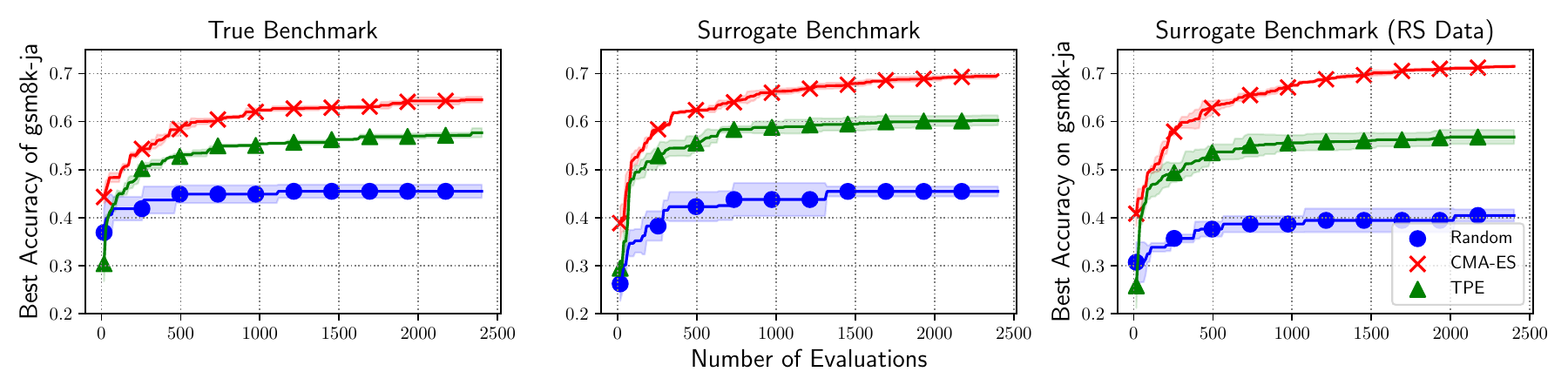}
    \vspace{-0.25cm}
    \caption{Search space for DFS merging (\texttt{SMM-Bench-DFS})}
    \label{fig:dfs_traj}
  \end{subfigure}
  \vspace{-0.2cm}
  \caption{Transitions of best accuracy of \texttt{gsm8k-ja} on true benchmark (left), surrogate benchmark (middle), and surrogate benchmark when using only a random sampled dataset (right). Random search, CMA-ES, and TPE are compared. The mean and standard deviation over ten and three runs are plotted in the search spaces for PS and DFS merging, respectively.}
  \label{fig:traj}
\end{figure}

We simulate optimization trajectories using \texttt{SMM-Bench}. Figures \ref{fig:ps_traj} and \ref{fig:dfs_traj} show the transitions of the best accuracy for three algorithms on the search spaces for PS and DFS merging, respectively. In these figures, we display the transitions on true benchmark (i.e., using actual evaluations of merged models) and surrogate benchmarks constructed using all dataset and only random sampling data. We observe that our surrogate benchmarks can capture the behavior of algorithms on true benchmarks, although the performance prediction on the search space for DFS merging tends to overestimate. In addition, our surrogate benchmarks work well even when using only random sampled data.

\section{Benchmark Demonstration Using SMM-Bench-PS}

\begin{wrapfigure}{r}[0mm]{0.38\linewidth}
\vspace{-1cm}
    \centering
    \includegraphics[width=0.9\linewidth]{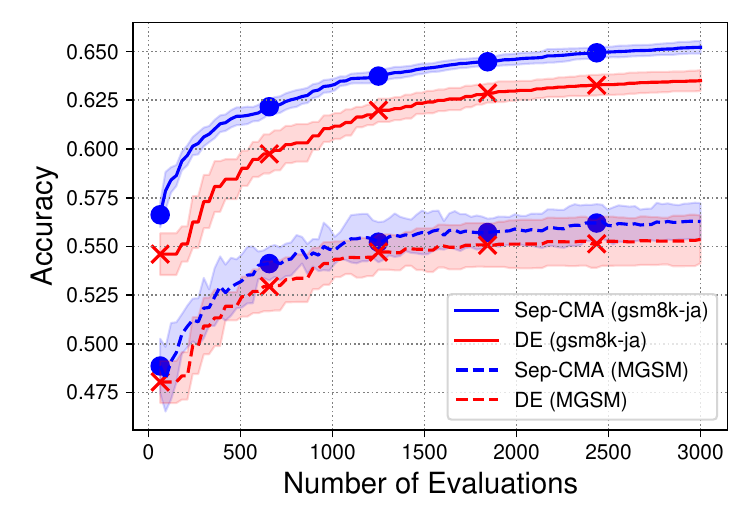}
    \vspace{-0.2cm}
  \caption{Performance of Sep-CMA and DE on \texttt{SMM-Bench-PS}.}
  \label{fig:demo}
\end{wrapfigure}

We ran two algorithms, separable CMA-ES (Sep-CMA)~\citep{Ros2008,nomura2024cmaes} with the default setting and differential evolution (DE)~\citep{Storn1997} implemented in \texttt{SciPy} with a population size of 64, on \texttt{SMM-Bench-PS}. Figure \ref{fig:demo} shows the transitions of accuracy for \texttt{gsm8k-ja} and \texttt{MGSM} datasets. We plotted the accuracy of the best solutions for \texttt{gsm8k-ja} that is the objective value in optimization. The accuracy of \texttt{MGSM} is the test performance of the best solutions. We observe that Sep-CMA outperforms DE on both \texttt{gsm8k-ja} and \texttt{MGSM} datasets, although the test performance difference is not so significant. This experimental comparison can be conducted in several minutes on a laptop owing to the surrogate benchmark. However, it requires many GPU days if it uses the actual evaluation of merged LLMs. Our surrogate benchmarks will be useful for hyperparameter tuning and comprehensive evaluation for optimizers.

\section{Conclusion}
We have proposed and evaluated surrogate benchmarks for model merging optimization. We also demonstrated a performance comparison of optimization algorithms not used for data collection on our benchmark. We believe that our surrogate benchmarks will contribute to algorithm development for model merging optimization and reproducible algorithm comparison.

\begin{acknowledgements}
We used ABCI 3.0 provided by AIST and AIST Solutions with support from ``ABCI 3.0 Development Acceleration Use.''
\end{acknowledgements}

\bibliography{mybib}

\end{document}